\def\year{2020}\relax
\documentclass[letterpaper]{article} 
\usepackage{aaai20}  
\usepackage{times}  
\usepackage{helvet} 
\usepackage{courier}  
\usepackage[hyphens]{url}  
\usepackage{graphicx} 
\usepackage{graphicx}  
\usepackage[utf8]{inputenc} 
\usepackage[T1]{fontenc}    
\usepackage{url}            
\usepackage{booktabs}       
\usepackage{amsfonts}       
\usepackage{nicefrac}       
\usepackage{microtype}      
\usepackage{times}
\usepackage{latexsym}
\usepackage{amsmath}
\usepackage{amstext}
\usepackage{graphicx}
\usepackage{ stmaryrd }
\usepackage{amssymb} 
\usepackage[caption=false]{subfig}
\usepackage{mathtools}
\usepackage{multirow}

\def\url#1{\expandafter\string\csname #1\endcsname}

\title{Harnessing GANs for Zero-Shot Learning of New Classes in Visual Speech Recognition}

\author{
Yaman Kumar\thanks{These authors \textit{contributed equally}. Author order was determined by a coin flip.} \\ Adobe \\ykumar@adobe.com
\And Dhruva Sahrawat\textsuperscript{$*$} \\ NUS, Singapore \\dhruva@comp.nus.edu.sg
\And Shubham Maheshwari\textsuperscript{$*$} \\ MIDAS Lab, IIIT-Delhi\\shubham14101@iiitd.ac.in
\And Debanjan Mahata \\Bloomberg, USA \\dmahata@bloomberg.net
\AND Amanda Stent \\Bloomberg, USA \\astent@bloomberg.net
\And Yifang Yin \\NUS, Singapore \\idsyin@nus.edu.sg 
\And Rajiv Ratn Shah \\MIDAS Lab, IIIT-Delhi\\rajivratn@iiitd.ac.in
\And Roger Zimmermann \\NUS, Singapore\\rogerz@comp.nus.edu.sg
}

\begin{document}

\maketitle

\begin{abstract}
Visual Speech Recognition (VSR) is the process of recognizing or interpreting speech by watching the lip movements of the speaker. 
%
Recent machine learning based approaches model VSR as a classification problem; however, the scarcity of training data leads to  error-prone systems with very low accuracies in predicting unseen classes. 
To solve this problem, we present a novel approach to zero-shot learning by generating new classes using Generative Adversarial Networks (GANs), and show how the addition of unseen class samples increases the accuracy of a VSR system by a significant margin of 27\% and allows it to handle speaker-independent out-of-vocabulary phrases.
We also show that our models are language agnostic and therefore  capable of seamlessly generating, using English training data, videos for a new language (Hindi). 
To the best of our knowledge, this is the first work to show empirical evidence of the use of GANs for generating training samples of unseen classes in the domain of VSR, hence facilitating zero-shot learning. We make the added videos for new classes publicly available along with our code\footnote{The code as well as the supplementary file can be found at https://github.com/midas-research/DECA/. The videos can be accessed through this YouTube playlist link: https://bit.ly/336geft. }.

\end{abstract}

\section{Introduction}

In Visual Speech Recognition (VSR), also known as automated lip reading, a system has to recognize words spoken by a human in a silent video, by primarily focusing on the region of the speaker's lips. VSR has a wide variety of potential applications such as in cybersecurity, and as assistive/augmentative technology  \cite{schmidt2013using}. For example, if a person has a laryngectomy or voice-box cancer, dysarthria, or works in very loud environments such as a factory floor, VSR can be enabling. 

VSR is treated as a classification problem. The training data can have single or multiple views of different speakers, paired with utterance labels (e.g. digits, phrases or sentences). Given a video displaying lip movements of a speaker, the model outputs a probability distribution over the utterance labels.
The final prediction is the utterance label with the highest probability score.
%
The performance of humans in lip reading is poor, with the best lip readers 
achieving an accuracy of just around 21\% \cite{easton1982perceptual}, so VSR is a challenging task.

Modeling VSR as a classification problem limits the predictions to a fixed number of utterances. 
This significantly hinders model generalization on predicting \textit{out-of-vocabulary} (OOV) classes, i.e. new utterances, new poses or new speakers. 
Additionally, preparing datasets for training VSR models is a challenging task by itself as it requires collecting hours of video recordings of utterances from multiple speakers and multiple poses. Recently, significant effort has gone into collecting high-quality VSR datasets, yet VSR is still both more complex than and has far less available training data than its close relative, ``classic" audio speech recognition. Therefore in functionality VSR is basically where audio speech recognition was in the late 1980s - isolated words/phrases, limited speakers.

Due to the unique aspects of VSR as a task, the challenges associated with it and its potential applications, it forms an appropriate down-stream task for low-shot OOV problem settings.
The challenge of automatically learning new classes and to generalize from very few labeled examples is often called \textit{low-shot} (or \textit{few-shot}) learning \cite{wang2018low}; or \textit{zero-shot} learning \cite{socher2013zero}, in situations where there are no examples of certain classes at all. Two main techniques applied to solve such machine learning problems are: (i) \textit{data driven approaches} \cite{krizhevsky2012imagenet}, which rely on adding more data either by collecting data from external sources or by augmenting existing datasets by synthesizing similar data points, and (ii)
\textit{parameter driven approaches} \cite{elhoseiny2017link}, which rely on regularization techniques in order to constrain the number of parameters to be learnt by the model.

In 2014, Goodfellow et al.  presented the idea of \textit{Generative Adversarial Networks} (GANs) \shortcite{goodfellow2014generative}.
Since then, there has been significant research on GANs and their applications \cite{creswell2018generative}.
In their seminal work, Goodfellow et al.  mention data augmentation as one of the applications of GANs, and indeed there is a growing literature focusing on it 
\cite{antoniou2017data,wang2018low}.
However, to date, the primary use of GANs for image/video data augmentation has been as a proxy to  traditional approaches such as rotation, reflection, cropping, translation, scaling and adding Gaussian noise \cite{krizhevsky2012imagenet}, and augmenting training data for a classification problem by generating similar data items
\cite{antoniou2017data}.
Recently, GANs have been used to generate unseen images from text descriptions
\cite{zhu2018generative}.

However, to the best of our knowledge, no prior work uses GANs to augment VSR datasets
with data items representing \textit{new classes} and performing zero-shot learning. 

\begin{figure*}[ht!]
    \centering
    \includegraphics[width=0.8\textwidth]{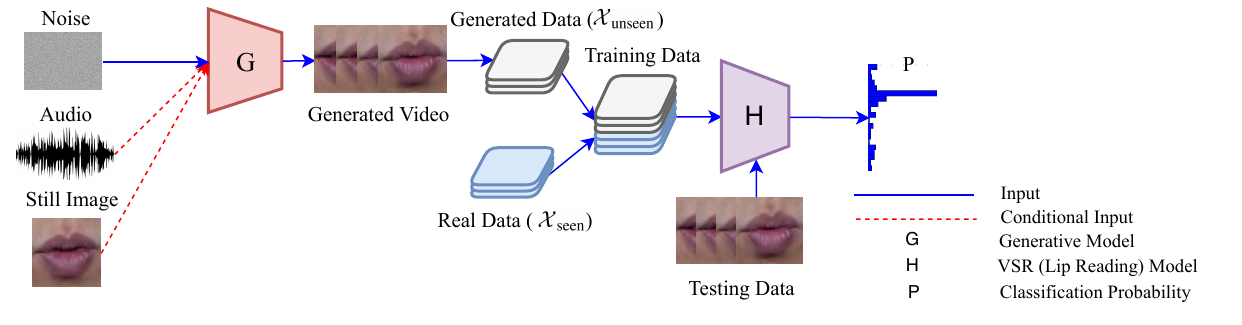}
    \caption{Overall pipeline demonstrated with the generator $G$ as the TC-GAN architecture}
    \label{fig:overal_pipeline}
\end{figure*}

In this paper, we describe how we generate lip movement videos of \textit{unseen} utterances using \textit{Temporal Conditional Generative Adversarial Networks} (TC-GANs) \cite{vougioukas2018end} and a \textit{viseme-concatenation} approach \cite{huang2002triphone} (Section \ref{sec:methodology}). As illustrated in Figure \ref{fig:overal_pipeline}, we  use the generated data for training a deep learning model for VSR. We empirically show the usefulness of the augmented data in increasing the accuracy of the VSR model while predicting unseen utterances. Additionally, we also show that our solution does not only cater to unseen utterances of the same language but also works well for a new  language. 

\noindent \textbf{Our Contributions -} 
The key contributions we make are summarized as follows:





\begin{itemize}
    \item We introduce zero-shot learning using GANs for augmenting VSR datasets with data for unseen utterances. 

\item We show that our solution results in improved accuracies of up to 27\% in prediction of unseen utterances at test time.

\item We show results for the cold-start problem (no training data at all for training the VSR model) on lipreading for the first-time in the literature \cite{amplayo2018cold}.

\item We further show that this method can work for zero-shot learning for an unseen language, Hindi.


\item We expand the OuluVS2 dataset \cite{anina2015ouluvs2} by adding videos for 10 new Hindi phrases and  25 new English phrases generated using our model\footnote{More details are present in supplementary materials}. We plan to release this dataset as well as the code for generating it. But due to space constraints, we only conduct experiments on 3 new Hindi phrases and the 10 phrases present in OuluVS2.

\end{itemize}

\section{Related Work}
\label{sec:related}

\subsection{Visual Speech Recognition}

Most early approaches to VSR such as \cite{ngiam2011multimodal} treat it as a single-word classification task. Other researchers such as  \cite{goldschen1997continuous} use methods borrowed from acoustic speech recognition like HMMs.  However, most work has focused on predicting utterances from limited lexicons. Furthermore, these techniques heavily rely upon hand-engineered feature pipelines. 

With advancements in deep learning and the increasing availability of large lip reading datasets, there have been approaches such as \cite{wand2016lipreading,zhou2019modality,salik2019lipper} addressing lip reading using deep learning based algorithms such as CNNs and LSTMs. Further, there have been works such as \cite{lee2016multi,kumar2018mylipper,Uttam2019,kumar2019lipper} extending lip reading from single view to multi-view settings by incorporating videos of mouth sections from multiple views together. Multi-view lip reading has been shown to improve performance significantly as compared to single view. Although the deep learning-based methods don't require hand-engineered features, most still treat lip reading as a classification task.

\subsection{Few- and Zero-Shot Learning}
%
In few-shot learning, one learns a model that can generalize to classes that have few examples in the training set \cite{lake2015human}. The few-shot learning problem has been studied from different perspectives such as optimization \cite{ravi2016optimization}, metric learning \cite{vinyals2016matching}, similarity-matching \cite{koch2015siamese}, and hierarchical graphical models \cite{salakhutdinov2012one}. However, generalizing a classifier on \textit{unseen} classes is mostly uninvestigated. 
We show that GANs can generate data for classes unseen during training, and that this generated data can give improved accuracy for the downstream VSR task. 
This problem falls under the meta-learning umbrella \cite{thrun1998lifelong,shrivastava2019mobivsr}.
High-level visual concepts like camera-pose or brightness, or even higher-level concepts like articulation (in the case of VSR) are shared across  computer vision problems.
Contemporary meta-learners, however, largely ignore these cross-task structures \cite{wang2018low}.
Using augmentation strategies like cropping and shifting, researchers have tried to make meta-leaners learn some of these high-level visual concepts.
However, high-level concepts like articulation require a different type of augmentation than simply changing brightness or camera angle; one needs a way to break down the high-level concept into decomposable elements \cite{zhu2018generative}. In the case of \cite{zhu2018generative}, the decomposable elements are keywords. In the case of VSR, the decomposable basic elements of articulation are \textit{visemes}, the visual equivalents of phonemes. 

Several research studies have shown that by knowing the probability distribution on transformations, one can generate examples from an unseen class by applying those transformations \cite{hariharan2017low,wang2018low,shah2017multimodal}.
Some authors \cite{hariharan2017low}, try to first learn transformations from some examples of a known class and then apply those transformations to some \textit{seed} examples of a new class.
This approach increases the data size of the novel class and hence helps in boosting accuracy.
In our case, the transformations are the conditions which we impose on temporal-conditional GANs to make them generate new visemes. We present these conditions and transformations in Section \ref{sec:methodology}.
%

%
%


\section{Dataset}
\label{dataset}

We use the OuluVS2 dataset \cite{anina2015ouluvs2}, a multi-view audio visual dataset. OuluVS2 has been used in several previous VSR studies \cite{chung2017lip,kumar2018mylipper,kumar2018harnessing}. It contains data from 52 speakers. Each speaker is filmed speaking three types of utterance: 
\begin{itemize}
    \item \textit{Oulu-digits} - digit sequences like \textit{1, 9, 0, 7, 8, 8, 0, 3, 2, 8}.
    \item \textit{Oulu-phrases} - ten short common English phrases such as \textit{``Excuse me''} and \textit{``see you''}.
    \item \textit{Oulu-sentences} - randomly chosen TIMIT sentences \cite{garofolo1993timit},  such as \textit{``Basketball can be an entertaining sport''} and \textit{``Are holiday aprons available to us?''}. 
\end{itemize}
Speakers are filmed from five different poses:  \{0$^\circ$,30$^\circ$,45$^\circ$,60$^\circ$,90$^\circ$\}.

We describe our training, testing, and validation splits in Section \ref{sec:exp}, along with the relevant experiments.

\section{Problem Formulation}
\label{problem-formulation}
 In this section, we describe  the data-driven approach we  followed in this paper.



 In general, a supervised classifier  can be depicted as learning a conditional distribution $p(y_{i}|x_{i})$, s.t. 
 $(x_{i}, y_{i})\in \mathcal{S}\textsubscript{train}$,  the set of training samples. The set $\mathcal{S}\textsubscript{test}$ is used for testing the  model. The set of all the labels present in $\mathcal{S}\textsubscript{train}$  is given by $\mathcal{Y}\textsubscript{train}$ (also referred as $\mathcal{Y}\textsubscript{seen}$ later in the text), and $\mathcal{Y}\textsubscript{test}$ gives the set of all the labels present in $\mathcal{S}\textsubscript{test}$. Similarly, $\mathcal{X}\textsubscript{train}$ represents the set of all the inputs present in $\mathcal{S}\textsubscript{train}$ (also referred as $\mathcal{X}\textsubscript{seen}$ later), and $\mathcal{X}\textsubscript{test}$ denotes those in $\mathcal{S}\textsubscript{test}$.

 In our problem setting, unseen classes are present in the test dataset i.e., $\mathcal{Y}\textsubscript{test}$ - $\mathcal{Y}\textsubscript{train}\neq\phi$. Let $\mathcal{Y}\textsubscript{test}$ - $\mathcal{Y}\textsubscript{train}$ = $\mathcal{Y}\textsubscript{unseen}$.
 Let the inputs corresponding to $\mathcal{Y}\textsubscript{unseen}$ be denoted as $\mathcal{X}\textsubscript{unseen}$. Thus, $\mathcal{Y}\textsubscript{unseen}$ represents all the \textit{unseen classes}, the classes which the model does not get to see while training.

 Our main objective in this work is to enable the model to correctly predict the labels present in $\mathcal{Y}\textsubscript{unseen}$, given that  $\mathcal{Y}\textsubscript{unseen}$ is available but no $x_{i} \in \mathcal{X}\textsubscript{unseen}$, is available for training. In other words, we aim to increase accuracy on $\mathcal{X}\textsubscript{unseen}$ using the knowledge available at train time ($\mathcal{S}\textsubscript{train}$, $\mathcal{Y}\textsubscript{unseen}$), while not decreasing accuracy on  $\mathcal{X}\textsubscript{train}$.
 
 To solve this problem, we ``hallucinate"  $\mathcal{X}\textsubscript{unseen}$ using $\mathcal{S}\textsubscript{train}$ and the labels present in  $\mathcal{Y}\textsubscript{unseen}$. We do this by learning a function $f: \mathcal{Y} \shortrightarrow \mathcal{X}$.
 For approximating $\mathcal{X}\textsubscript{unseen}$ with $f$ using machine learning, we only have  $\mathcal{S}\textsubscript{train}$ which, notably does not have any information about $\mathcal{X}\textsubscript{unseen}$.
 For solving this problem, we introduce an intermediate representation $\mathcal{Z}$ such that $g: \mathcal{Y} \shortrightarrow \mathcal{Z}$ and $h: \mathcal{Z} \shortrightarrow \mathcal{X}$.

 In the case of lipreading, $\mathcal{X}$ represents videos of speakers and $\mathcal{Y}$ represents the utterance labels. 
 $\mathcal{Z}$ represents the \textit{audio} representation of the utterances. The function $g$ takes an utterance label and converts it into its corresponding audio. The function $h$ is approximated in this work by a TC-GAN model or a viseme-concatenation approach. The function $h$ takes in audio as input and produces a corresponding video as output. The input-output relation between audio and video is not one-to-one since videos of different speakers in different poses have the same audio. In order to make this mapping one-to-one, we introduce speaker and pose information $\mathcal{I}$; $h$ becomes $h: \mathcal{Z}\times\mathcal{I} \shortrightarrow \mathcal{X}$. The pair of (audio, speaker~image~in~the~specified~pose~angle) uniquely identifies a video.


 The overall data hallucination pipeline is as follows:
 \begin{itemize}
     
     \item The utterances present in the set  $\mathcal{Y}\textsubscript{unseen}$ are projected onto the audio space $\mathcal{Z}$ (Sections \ref{sec:methodology} and \ref{sec:exp}).
    
     \item 
     A speaker's image chosen randomly from the training subset of \textit{Oulu-phrases} and the audio projections of utterance $z_{i}$ (where $z_{i}\in \mathcal{Z} $) help the generator models (\textit{i.e.} TC-GAN and viseme-concatenation)  generate videos for the set $\mathcal{X}^{TC-GAN}_{unseen}$ and $\mathcal{X}_{unseen}^{viseme-concat}$ (Section \ref{sec:methodology}).

     \item The sets $\mathcal{X}_{unseen}^{TC-GAN}$ and $\mathcal{X}_{unseen}^{viseme-concat}$ are then included individually in  $\mathcal{S}\textsubscript{train}$ in different experiments. These different experiments allow us to gauge the efficacy of the two approaches of generating unseen videos separately (Section \ref{sec:exp}).
    
     \end{itemize}
     
    The model thus trained is then tested on the full $\mathcal{S}\textsubscript{test}$ which contains original (rather than hallucinated) videos of both seen and unseen classes (Section \ref{sec:exp}).

A question that generally gets asked is how the viseme-concatenation and TC-GAN approaches are able to generate videos of unseen classes just by taking audio as input. The viseme-concatenation approach uses a manual mapping of phonemes to visemes. The TC-GAN approach, by contrast, learns to maps phonemes to visemes. The TC-GAN approach \textit{also} learns inter-visemic frames, \textit{i.e.}, the frames which occur between two visemes. These are intermediate mouth formations that stitch together articulations. The importance of these inter-visemic frames can be judged by the fact that the viseme-concatenation approach does not result in as good a performance as the TC-GAN approach (Section \ref{sec:exp}). 


\section{Methodology}
\label{sec:methodology}
In this paper, we apply two techniques for zero-shot learning to VSR; first, a Generative Adversarial Network (GAN) approach, and second,   a viseme-concatenation approach. Both approaches synthesize examples belonging to classes unseen at training time ($\mathcal{Y}\textsubscript{unseen}$). The overall process is depicted in Figure \ref{fig:overal_pipeline}. Our pipeline is primarily composed of two components - a \textit{generator} \textbf{G} (Sections \ref{ssec:generative-model} and \ref{ssec:concatenationgen}), and a \textit{classifier network} \textbf{H}, whose main objective is to perform the task of VSR as described in Section \ref{ssec:alr}. \textbf{G}'s task is to hallucinate lip movement videos for utterances 
not present in
the training data ($\mathcal{Y}\textsubscript{unseen}$), 
as if they were uttered by a given speaker in a given pose. \textbf{H} is then trained on the videos generated by \textbf{G} as well as on real videos, hopefully making it a robust model which performs well on both seen and unseen class labels ($\mathcal{Y}\textsubscript{unseen}\cup\mathcal{Y}\textsubscript{seen}$). Next, we describe the individual components of our pipeline.  









\subsection{Temporal Conditional GAN}
\label{ssec:generative-model}

\begin{figure}[htbp]
    \centering
    \includegraphics[width=\columnwidth]{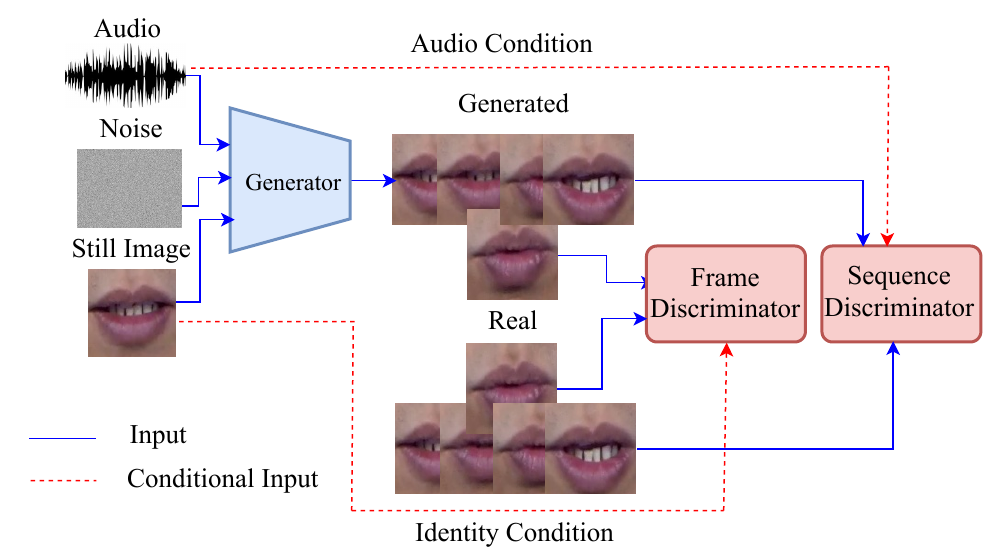}
    \caption{\textbf{Temporal Conditional GAN Architecture}}
    \label{fig:tc-gan}
\end{figure}

Our TC-GAN model relies heavily on the recently published work of Vougioukas, Petridis and Pantic \shortcite{vougioukas2018end}, who generated lip movement animations using \textit{Temporal Conditional GANs} (TC-GANs). We reimplemented the model from this paper, and trained it on \textit{Oulu-sentences}. The TC-GAN is composed of three main components as shown in Figure \ref{fig:tc-gan}, comprising a single \textit{generator}, and two discriminators - (a) a \textit{frame discriminator}, and (b) a \textit{sequence discriminator}. Here, we briefly describe these components\footnote{Please refer to the original work \cite{vougioukas2018end} for a detailed description and also refer to the supplementary materials to get a detailed list of the we parameters used in training our TC-GAN model.}.

\noindent \textbf{Generator}: The Generator takes three inputs: a) Gaussian noise, b) a still image of a speaker's lips from a particular pose angle, and c) audio of the utterance being spoken. The Generator is conditioned on the audio and the still image. The image encoding component of the popular segmentation algorithm U-Net \cite{ronneberger2015u} is used to encode the identity of the still image into a  latent vector ($L_I$). The audio file is divided into small chunks of overlapping samples, passed through a 1-D CNN, and then fed to a 2-layer GRU to generate an audio latent vector ($L_A$) for each corresponding time step. A Gaussian noise vector is generated and passed through a 1-Layer GRU at each time step producing the noise latent vector ($L_N$). The three latent vectors are concatenated to produce a single latent vector which is then passed through the decoder component of U-Net to produce the generated frame at that time step. The process is repeated temporally to generate animated frames corresponding to each  chunk of the audio sample.

\noindent \textbf{Frame Discriminator}: 
The Frame Discriminator is a 6-Layer CNN. It takes in an input frame, which can be real or hallucinated, concatenated channel-wise with the still image of the speaker's lips (the same still image as the one fed to the Generator) and is trained to discriminate whether the input frame is real or hallucinated.
  
\noindent \textbf{Sequence Discriminator}: The real or hallucinated video segment at each time step is passed through a 2-Layer GRU and encoded into a latent vector. The same is done for the audio sample at each time step. The two latent vectors produced are then concatenated and passed through a 2-Layer CNN that is trained to discriminate between real and hallucinated videos. 
\begin{multline} \label{adv_loss} 
    \mathcal{L}_{a d v}\left(D_{i m g}, D_{S e q}, G\right)= \mathbf{E}_{x \sim P_{d}}[\log D_{i m g}(S(x), x_{1})]+\\ \mathbf{E}_{z \sim P_{z}}\left[\log \left(1-D_{i m g}\left(S(G(z)), x_{1}\right)\right)\right]+\\ \mathbf{E}_{x \sim P_{d}}\left[\log D_{s e q}(x, a)\right]+\\ \mathbf{E}_{z \sim P_{z}}\left[\log \left(1-D_{s e q}(G(z), a)\right)\right] 
\end{multline}
Equation \ref{adv_loss} represents the adversarial loss function of the model where $D_{i m g}$ is the frame discriminator, $D_{S e q}$ is the sequence discriminator, $G$ is the frame generator, $x_{1}$ and $a$ are the still image and audio used as conditional input, and $x$ is the sequence of frames in the real video.
\begin{equation}\label{l1_loss}
\mbox{\fontsize{10}{12}\selectfont\( %
\mathcal{L}_{L_{1}}=\sum_{p \in[0, W] \times\left[0,H\right]}\left|F_{p}-G_{p}\right|  
\)}
\end{equation}
The pixel-wise $L_{1}$ reconstruction loss function on each real frame $F$ and generated frame $G$, both of dimension $W$ x $H$, is represented by Equation \ref{l1_loss}.

\begin{figure*}[htbp]
\centering

\subfloat[]
{
\includegraphics[width=0.75\columnwidth]{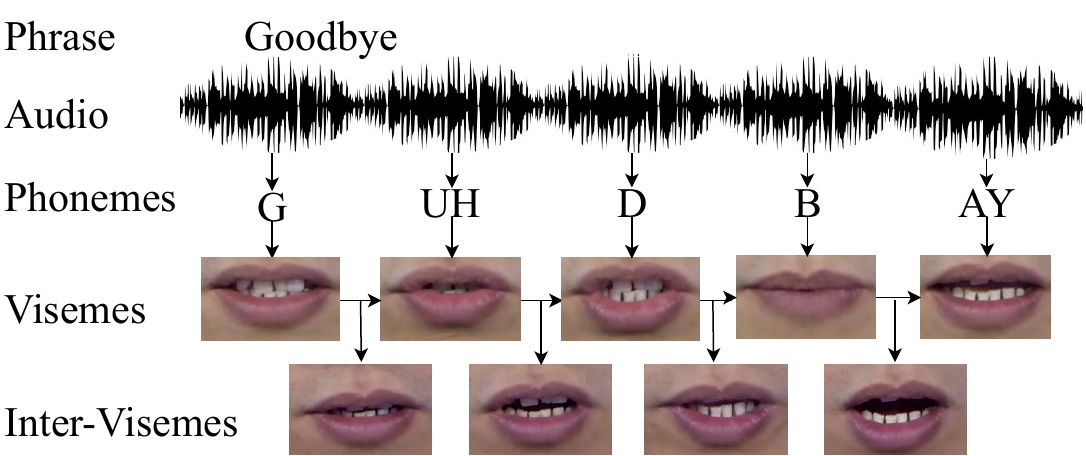} 
\label{fig:engviseme}
} \hspace{6mm}
\subfloat[]{\includegraphics[width=1\columnwidth]{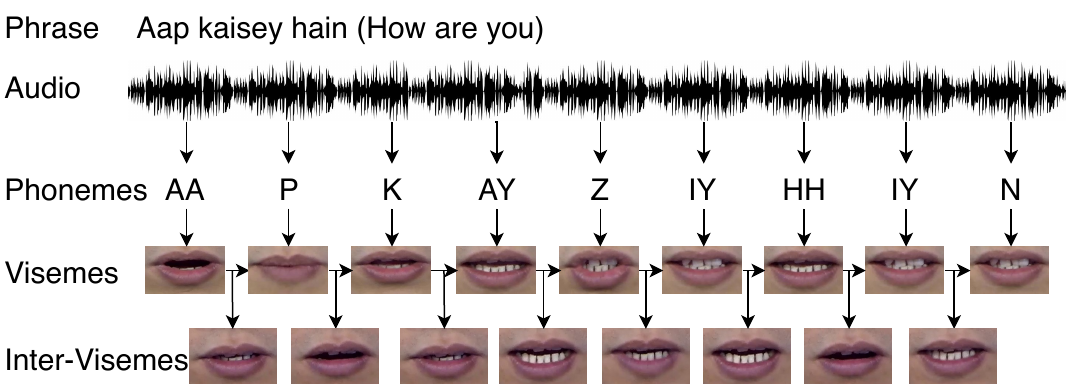}\label{fig:hindiviseme}} 
\\
\noindent 
\subfloat[]{
\includegraphics[width=0.75\columnwidth]{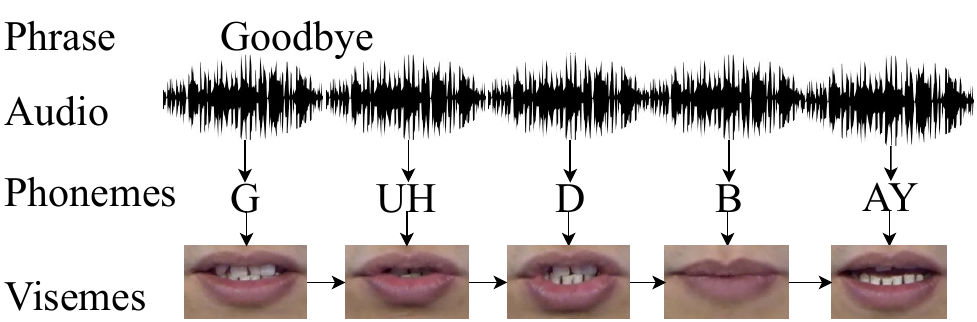}
\label{fig:engviseme_only}
} \hspace{6mm}%
\subfloat[]{\includegraphics[width=1\columnwidth]{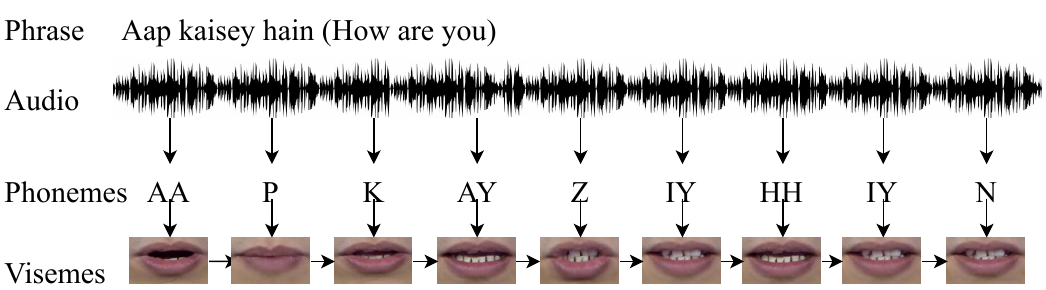}\label{fig:hindiviseme_only}}
\caption[caption]{TC-GAN output for an English phrase (a) and a Hindi phrase (b). Viseme-concatenation output for an English phrase (c) and a Hindi phrase (d).} 

\end{figure*}










For generating a video of a particular speaker speaking a given utterance, we follow these steps:
First, if the phrase is present in \textit{Oulu-phrases} then the audio file present in the dataset are used; otherwise, an off-the-shelf Text-To-Speech (TTS) software\footnote{https://ttsmp3.com} is used for generating the audio file. The audio file is sub-sampled into small overlapping chunks. Each chunk, along with the still image of the speaker, is given as input to the generative part of our model, which then outputs a video frame (an image) corresponding to that chunk. The resulting sequence of video frames is then concatenated to obtain the generated video, as shown in Figure \ref{fig:overal_pipeline}.


In essence, the model learns how to map phonemes to visemes. Our model is trained on \textit{Oulu-sentences}, which contains phonetically rich TIMIT sentences in the English language. Therefore, our model sees a rich set of phonemes for the English language during training. Based on this, we assume that it can map the majority of the phonemes present in the English language into their corresponding visemes.  
However, our model is not limited to only a fixed set of visemes; it also generates inter-visemic frames, as shown in Figure \ref{fig:engviseme}. These inter-visemic frames cannot be mapped to a distinct viseme; they are the images of intermediate mouth movements present between two different visemes. The generation of inter-visemic frames by the TC-GAN model ensures smooth and continuous mouth movements in the entire generated video. If the inter-visemic frames are removed, anyone viewing the video would not be able to understand what is being spoken and the whole video would appear unnatural and discontinuous. 

We also generate utterances in the Hindi language, which is an entirely different language from that which the model is trained on. The generation process of an example Hindi phrase \textit{Aap kaisey hain} (How are you) is shown in Figure \ref{fig:hindiviseme}. The phonemes present in the Hindi phrase are [`AA', `P', `K', `AY',`Z',`IY',`HH',`IY',`N'], and are also present in the TIMIT sentences. Since the phonemes are common and our model has learned to map phonemes to visemes, it can also generate a video of a speaker speaking the given Hindi utterance.

\subsection{Concatenation of Visemic Frames}
\label{ssec:concatenationgen}


The other technique we use for data generation is a Concatenative Visual Speech Synthesis technique \cite{huang2002triphone}, which works by using visemes to generate video. Each viseme is just an image of a mouth movement of a speaker corresponding to a particular sound. In this technique, video of lip movements for a speaker speaking a given utterance is generated by selecting appropriate viseme frames from a set of visemes belonging to that speaker and then concatenating them. This technique requires a speaker-to-viseme database which contains the set of all the visemes for every speaker whose video we want to generate. We construct this database \footnote{It is present in the supplementary file.} by parsing through all the videos present in  \textit{Oulu-sentences} and manually annotating the set of visemes for every speaker present in the dataset.

An example of the generation process using this technique for a speaker speaking the English utterance \textit{Goodbye} is shown in Figure \ref{fig:engviseme_only}. For generating a video of a speaker speaking the given English utterance, first, we convert the utterance into a sequence of phonemes using the CMU Pronouncing Dictionary \footnote{http://www.speech.cs.cmu.edu/cgi-bin/cmudict}. The resulting sequence of phonemes  [`G',`UH',`D',`B',`AY'] is then matched to a sequence of visemes. This is done by using  the Jeffers phoneme-to-viseme map \cite{phone2viseme_jeff}\footnote{More details are present in the supplementary materials.}, and our speaker-to-viseme database. Note that the \textit{Oulu-sentences} segment of the dataset contains all the phonemes present in the English language. Therefore it also contains all the visemes which can be mapped from any phoneme present in the English language. The resulting sequence of visemes is then concatenated to obtain the generated video.

We also generate phrases in the Hindi language. An example is the Hindi phrase \textit{Aap kaisey hain} (How are you) shown in Figure \ref{fig:hindiviseme_only}. Since the CMU Pronouncing Dictionary does not work for Hindi phrases, we use \textit{phonemizer}\footnote{https://github.com/bootphon/phonemizer}, which is a multi-lingual text-to-phoneme converter, for converting an input Hindi phrase into a sequence of phonemes. Since most of  the phonemes present in Hindi are also present in \textit{Oulu-sentences}, we can use the Jeffers phoneme-to-viseme map and our speaker-to-viseme database to map the sequence of phonemes into a sequence of visemes. The resulting sequence of visemes is then concatenated to obtain the generated video.

\subsection{VSR Model}
\label{ssec:alr}

For VSR, we implement and train the architecture from \cite{petridis2017enda}. The architecture consists of three parts: an encoder, two BiLSTM layers, and a fully connected layer. The encoder is composed of 3 fully connected layers. The resultant encoded information is concatenated with its first and second derivative and fed into the first BiLSTM layer. The first BiLSTM layer captures the temporal information associated with each view, and the second BiLSTM layer is used to fuse the information from different views. The last fully-connected layer produces the final output class probabilities. We use PyTorch \cite{paszke2017automatic}, and release our implementation  as part of the supplementary material.






\section{Experiments}
\label{sec:exp}

    


\subsection{Experimental Setting}
\label{ssection:trainconfig}

In this section, we present experimental results:
 \begin{itemize}

\item comparing the TC-GAN and viseme-concatenation approaches for adding new (unseen) classes to  $\mathcal{Y}\textsubscript{unseen}$ (Section \ref{ssec:set-1}).

\item assessing the extent to which a VSR model can harness hallucinated data for improving accuracy on $\mathcal{Y}\textsubscript{unseen}$ while maintaining accuracy for  $\mathcal{Y}\textsubscript{seen}$.

\item evaluating the ability of the overall system to adapt to a new language (Section~\ref{ssec:set-2}).

\end{itemize}

In order to conduct our experiments we train three different model configurations:
\begin{itemize}
    \item \textbf{Model 1} - a model trained only on \textit{Oulu-phrases}.
    \item \textbf{Model 2} - a model trained on \textit{Oulu-phrases} and on videos generated using the TC-GAN model.
    \item \textbf{Model 3} - a model trained on \textit{Oulu-phrases} and on videos generated using viseme-concatenation.
\end{itemize}

\begin{table}[ht!]
\centering
\caption{ Overall top-3 accuracy (\%) for each model on $\mathcal{Y}\textsubscript{unseen} \cup \mathcal{Y}\textsubscript{seen}$.
}

\label{rtable:1}
\resizebox{1\columnwidth}{!}{%
\begin{tabular}{lllllll}
\hline
\vtop{\hbox{\strut \textbf{S.}}\hbox{\strut \textbf{No.}}} & \vtop{\hbox{\strut \textbf{Seen}}\hbox{\strut \textbf{classes}}}
 & \vtop{\hbox{\strut \textbf{Unseen}}\hbox{\strut \textbf{classes}}} & \textbf{Model 1} & \textbf{Model 2} & \textbf{Model 3} \\ \hline
\textbf{1} & 9 & 1  & 90   & 98.3  & 86.4 \\ \hline
\textbf{2} & 8 & 2  & 79.4 & 90    & 76.4 \\ \hline
\textbf{3} & 7 & 3  & 70   & 84.4  & 67.8 \\ \hline
\textbf{4} & 6 & 4  & 60   & 72.2  & 60.5 \\ \hline
\textbf{5} & 0 & 10 & 30.8 & 70    & 34.7 \\ \hline
\end{tabular}
}
\end{table}

\begin{table}[ht]
\centering
\caption{Overall top-1 accuracy (\%) for each model on $\mathcal{Y}\textsubscript{unseen} \cup \mathcal{Y}\textsubscript{seen}$. }

\label{rtable:acc_1}
\resizebox{\columnwidth}{!}{%
\begin{tabular}{lllllll}
\hline
\vtop{\hbox{\strut \textbf{S.}}\hbox{\strut \textbf{No.}}} & \vtop{\hbox{\strut \textbf{Seen}}\hbox{\strut \textbf{classes}}}
 & \vtop{\hbox{\strut \textbf{Unseen}}\hbox{\strut \textbf{classes}}} & \textbf{Model 1} & \textbf{Model 2} & \textbf{Model 3} \\ \hline
\textbf{1} & 9 & 1  & 87.2 & 89.7 & 81.1 \\ \hline
\textbf{2} & 8 & 2  & 77.2 & 81.4 & 70.8 \\ \hline
\textbf{3} & 7 & 3  & 69.2 & 72.8 & 62.8 \\ \hline
\textbf{4} & 6 & 4  & 58.9 & 65.3 & 54.7 \\ \hline
\textbf{5} & 0 & 10 & 13.3 & 40   & 17.5 \\ \hline
\end{tabular}
}
\end{table}

As suggested by \cite{petridis2017enda,anina2015ouluvs2,kumar2018mylipper}, we take 35 speakers for training, 5 speakers for validation, and the remaining 12 speakers for testing the different models. For \textit{Oulu-phrases}, which is used to train the VSR model, this results in a total of 5250, 750, and 1800 videos belonging to train, test, and validation data respectively.
Similarly, for \textit{Oulu-sentences}, which is used to train the TC-GAN model, this results in a total of 875, 125, and 300 videos belonging to train, test, and validation data respectively. It is to be noted that the VSR model and the data generation models share the same training, validation and test speaker splits, ensuring that no data belonging to the test speakers is seen by the data generating models during training and vice versa.


In our experiments, Model 1 is our baseline. The major differences between Models 2 and 3 are that the videos belonging to Model 2, on account of being generated using TC-GANs, have inter-visemic frames. Thus, Model 2's videos have articulatory continuity while frames in videos for Model 3 are non-continuous.

We report top-1 and top-3 accuracy. Top-3 accuracy denotes whether the true utterance occurs in the top-3 predicted utterances ranked by confidence. Top-1 and Top-3 results follow the same trend. 

Note that since our dataset has multiple views (pose angles), we have a separate TC-GAN model for each view and the pose of the input speaker lip region image corresponds with the TC-GAN model used. For handling different view videos as input, we have a different VSR model corresponding to each view combination. During evaluation, we report the results on the test split for the view combination VSR model which has the highest score on the validation split. For all the different sets of experiments, the best view-combination is given in Table \ref{rtable:view_combination}.

\begin{table}[]
\centering
\caption{The best view combination ranked according to their score on validation split.}
\label{rtable:view_combination}
\resizebox{1\columnwidth}{!}{%
\begin{tabular}{ccccc}
\hline
\vtop{\hbox{\strut \textbf{Exper-}}\hbox{\strut \textbf{iment}}}  & \vtop{\hbox{\strut \textbf{S.}}\hbox{\strut \textbf{No.}}} & \vtop{\hbox{\strut \textbf{Model}}\hbox{\strut \textbf{\hspace{2.5mm}   1}}} &  \vtop{\hbox{\strut \textbf{Model}}\hbox{\strut \textbf{\hspace{2.5mm}   2}}} & \vtop{\hbox{\strut \textbf{Model}}\hbox{\strut \textbf{\hspace{2.5mm}   3}}}\\ \hline
\multirow{5}{*}{\textbf{Set-1}} & \textbf{1}  & \{0$^\circ$,30$^\circ$,45$^\circ$ ,60$^\circ$,90$^\circ$\}      & \{30$^\circ$,60$^\circ$,90$^\circ$\}  			& \{0$^\circ$\}  \\ \cline{2-5} 
                                & \textbf{2}  & \{0$^\circ$,45$^\circ$ ,90$^\circ$\}                            & \{0$^\circ$,45$^\circ$ ,60$^\circ$,90$^\circ$\}   						& \{0$^\circ$\}  \\ \cline{2-5} 
                                & \textbf{3}  & \{0$^\circ$,60$^\circ$,90$^\circ$\}                             & \{45$^\circ$ ,90$^\circ$\}    		& \{0$^\circ$\}   \\ \cline{2-5} 
                                & \textbf{4}  & \{0$^\circ$,30$^\circ$,60$^\circ$,90$^\circ$\}                  & \{45$^\circ$ ,60$^\circ$,90$^\circ$\}  				& \{0$^\circ$\}  \\ \cline{2-5} 
                                & \textbf{5}  & \{0$^\circ$\}                                                   & \{0$^\circ$,30$^\circ$,45$^\circ$ ,60$^\circ$\} 						& \{0$^\circ$\}  \\ \hline
\textbf{Set-2}                  & \textbf{1}  & \{45$^\circ$ ,60$^\circ$,90$^\circ$\}                           & \{0$^\circ$,30$^\circ$,45$^\circ$ ,90$^\circ$\}   						& \{0$^\circ$\}   \\ \hline
\end{tabular}
}
\end{table}


\subsection{Experiment-Set 1 : Experiments on Unseen Classes}
\label{ssec:set-1}
In this set of experiments, we test our models on \textit{unseen classes} in the same language  ($\mathcal{Y}\textsubscript{unseen}$). The unseen classes are utterances not present in the real training data. Model 1 therefore contains no instances of these utterances. Model 2 contains instances of these utterances hallucinated by the TC-GAN model, while model 3 contains instances of these utterances hallucinated by viseme-concatenation.

\begin{table}[ht]
\centering
\caption{Overall top-3 accuracy (\%) of each model on $\mathcal{Y}\textsubscript{unseen}$ and separately on $\mathcal{Y}\textsubscript{seen}$ (separated by /).
}
\label{rtable:2}
\begin{tabular*}{\columnwidth}{l@{\extracolsep{\fill}}ccc}
\hline
\textbf{Unseen classes} & \textbf{Model 1} & \textbf{Model 2} & \textbf{Model 3} \\ \hline
\textbf{1}     & 0  /  100                & 86.1  /  99.7             & 5.5  /  95.4             \\ \hline
\textbf{2}     & 0  /  99.3               & 52.8  /  99.3             & 1.4  /  95.1             \\ \hline
\textbf{3}     & 0  /  100                & 49.1  /  99.6             & 0.0  /  96.8             \\ \hline
\textbf{4}     & 0  /  100                & 31.2  /  99.5             & 8.3  /  95.4             \\ \hline
\end{tabular*}
\end{table}

\begin{table}[ht]
\centering
\caption{Overall top-1 accuracy (\%) of each model on $\mathcal{Y}\textsubscript{unseen}$ and separately on $\mathcal{Y}\textsubscript{seen}$ (separated by /). }
\label{rtable:acc_2}
\begin{tabular*}{\columnwidth}{l@{\extracolsep{\fill}}ccc}
\hline
\textbf{Unseen classes} & \textbf{Model 1} & \textbf{Model 2} & \textbf{Model 3} \\ \hline
\textbf{1}     & 0 / 96.9 & 33.3 / 96    & 0   / 90.1 \\ \hline
\textbf{2}     & 0 / 96.5 & 23.6 / 95.8  & 0   / 88.5 \\ \hline
\textbf{3}     & 0 / 98.8 & 21.3 / 94.8  & 0   / 89.7 \\ \hline
\textbf{4}     & 0 / 98.1 & 16   / 98.1  & 0.7 / 90.7 \\ \hline
\end{tabular*}
\end{table}

The top-3 and top-1 accuracies of each model on $\mathcal{Y}\textsubscript{seen}\cup\mathcal{Y}\textsubscript{unseen}$ are given in Tables \ref{rtable:1} and \ref{rtable:acc_1}. To make sure that training on new data for $\mathcal{Y}\textsubscript{unseen}$ does not negatively impact the accuracy for  $\mathcal{Y}\textsubscript{seen}$, we also report top-3 and top-1 accuracies on $\mathcal{Y}\textsubscript{unseen}$ and $\mathcal{Y}\textsubscript{seen}$ separately in Tables \ref{rtable:2} and \ref{rtable:acc_2}. 

We ran multiple experiments varying the number of unseen classes from 10\% to 40\%. Model 2 always performs better than Models 1 and 3 on unseen classes. The top-3 accuracy cost for Model 2 on seen classes is 0\%-0.5\%, depending on the number of unseen classes. Model 3 performs far worse than Model 2 on unseen classes, while incurring a bigger accuracy cost on seen classes.

As an extreme case, we also show the performance of the models when 100\% of the classes are unseen, in the last row in  Tables \ref{rtable:1} and \ref{rtable:acc_1}. This represents the cold-start problem for VSR systems \cite{amplayo2018cold}. On this type of problem, Model 1 gets a top-3 accuracy of just 30.8\% (an accuracy which is still better than randomly predicting any class) while Model 2 gets more than double that accuracy. This indicates that data hallucination can be used for cold-start for VSR.

\subsection{Experiment-Set 2 : Experiments on New Language}
\label{ssec:set-2}
In this set of experiments, our unseen classes are classes from a different language, Hindi. The utterances we use for this analysis are \{
\textit{Aap kaisey hai} (How are you), \textit{Kyaa chal rahaa hai} (What is going on), \textit{Shubh raatri} (Good night)
\}. We test the models by retaining English Oulu-phrases $\mathcal{Y}\textsubscript{seen}$ and introducing Hindi phrases as $\mathcal{Y}\textsubscript{unseen}$. The results on the test split pertaining to this setting are given in Table \ref{rtable:3}.

Both Model 2 and Model 3 perform better than the Model 1. This is expected since Model 1 has never seen any Hindi phrase. Model 2 performs better than Model 3 by a margin of 22.5\%. It is also interesting to note that Model 2 has a 100\% top-3 accuracy on the newly introduced Hindi phrases ($\mathcal{Y}\textsubscript{unseen}$), while Model 3 only has a 32.4\% top-3 accuracy.  Additionally, the top-1 accuracy of Model 2 and Model 1 is almost the same when evaluated on the seen classes, while we see a  drop in top-1 accuracy of more than 2\% in the case of Model 3. Both of these results again demonstrate the importance of the presence of inter-visemic frames in the generated video.

\begin{table}[ht]
\centering
\caption{Overall accuracy of each model on unseen classes from a different language (Hindi) 
}
\resizebox{1\columnwidth}{!}{%
\label{rtable:3}
\begin{tabular}{llll}
\hline
\textbf{Metrics}             & \textbf{Model 1} & \textbf{Model 2} & \textbf{Model 3} \\ \hline
\textbf{Top-1 Accuracy (Acc)}       & 72.6             & 95.1             & 72.6             \\ \hline
\textbf{Top-3 Acc}           & 76.7             & 99.6             & 83.9             \\ \hline
\textbf{Unseen Class Top-1 Acc}       & 0                & 96.3             & 7.4              \\ \hline
\textbf{Unseen Class Top-3 Acc} & 0                & 100              & 32.4             \\ \hline
\textbf{Seen Class Top-1 Acc}       & 94.4             & 94.7             & 92.2             \\ \hline
\textbf{Seen Class Top-3 Acc} & 99.7             & 99.4             & 99.4             \\ \hline
\end{tabular}
}
\end{table}

Since, for Model 2, new data has a negligible affect on seen class accuracy, and the difference between Model 1 and Model 2 is only the addition of the data generated by the TC-GAN model for unseen classes, we can safely assume that we can extend the number of classes significantly without adversely affecting the performance on seen classes. These results clearly show that Model 2 is able to clearly distinguish between Hindi and English phrases and is able to predict both with a very high degree of certainty compared to both Model 1 and Model 3.

\section{Conclusions and Future Work}
In this paper, we introduced a GAN-based method to deal with the zero-shot learning problem for VSR. We demonstrated that this method is more effective than viseme-concatenation for data augmentation for unseen classes, and that it does not cause appreciable degredation in accuracy for seen classes. We show that this method can be used to help a VSR model generalize to unseen classes and to a new language.

Although in this work, we explored the phonetically close language pair of English and Hindi, it would be interesting to explore it on phonetically-far language pairs, like English and Hopi, where new visemes must also be hallucinated. Additionally, we would like to explore TC-GAN data augmentation for continuous VSR, as opposed to the classification-based VSR approach we used in this paper.

\section*{Acknowledgement}
This research has been supported in part by Singapore Ministry of Education Academic Research Fund Tier 2 under MOE's official grant number MOE2018-T2-1-103.

Rajiv Ratn Shah is partly supported by the Infosys Center for AI, IIIT Delhi and ECRA Grant (ECR/2018/002776) by SERB, Government of India.


\fontsize{9.0pt}{10.0pt}
\selectfont 
\bibliographystyle{aaai}
\bibliography{references}

\end{document}